# Detecting Adversarial Perturbations Through Spatial Behavior in Activation Spaces

Ziv Katzir
Department of Software and Information Systems Engineering, Ben-Gurion University of the Negev
zivka@post.bgu.ac.il

Yuval Elovici
Department of Software and Information Systems Engineering, Ben-Gurion University of the Negev
elovici@bgu.ac.il

*Abstract*— Neural network based classifiers are still prone to manipulation through adversarial perturbations. State of the art attacks can overcome most of the defense or detection mechanisms suggested so far, and adversaries have the upper hand in this arms race.

Adversarial examples are designed to resemble the normal input from which they were constructed, while triggering an incorrect classification. This basic design goal leads to a characteristic spatial behavior within the context of Activation Spaces, a term coined by the authors to refer to the hyperspaces formed by the activation values of the network's layers. Within the output of the first layers of the network, an adversarial example is likely to resemble normal instances of the source class, while in the final layers such examples will diverge towards the adversary's target class. The steps below enable us to leverage this inherent "shift" from one class to another in order to form a novel adversarial example detector. We construct Euclidian spaces out of the activation values of each of the deep neural network layers. Then, we induce a set of k-nearest neighbor classifiers (k-NN), one per activation space of each neural network layer, using the non-adversarial examples. We leverage those classifiers to produce a sequence of class labels for each nonperturbed input sample and estimate the a priori probability for a class label change between one activation space and another. Training our detector with only normal input follows the principles of anomaly detection, aiming to make our method future ready for as yet unknown attack methods. During the detection phase we compute a sequence of classification labels for each input using the trained classifiers. We then estimate the likelihood of those classification sequences and show that adversarial sequences are far less likely than normal ones.

We evaluated our detection method against the state of the art C&W attack method, using two image classification datasets (MNIST, CIFAR-10) which are commonly used for adversarial evaluation. Our evaluation results show that our detector achieves an AUC of 0.95 for the CIFAR-10 dataset, with only a marginal increase in the computational complexity.

## I. INTRODUCTION

State of the art neural network based classifiers outperform humans in a wide range of tasks [31], [33]. However, they remain extremely vulnerable to adversarial manipulation. Following the initial discovery of adversarial examples in deep neural networks [32] an array of increasingly more powerful attack methods have been suggested [6], [25]. Adversarial perturbations were refined to the point that a single pixel can control the output of a convolutional neural network (CNN) based classifier [30], and recent research demonstrated how generic perturbations can be applied as is to a wide range of inputs and network architectures [19].

Furthermore [23], coined the term adversarial transferability, demonstrating that it is possible to launch an attack against a given network using perturbation patterns devised on a surrogate network. This discovery opened the door to black box attacks. Finally, the applicability of adversarial examples in the real world was shown in various studies [2], [15], [28].

Naturally, significant research efforts have been invested to try to counter such attacks, either by increasing classifier resilience [12], [18], [20], [26], [27] or attempting to differentiate adversarial examples from normal instances [9], [11], [29]. In what has evolved into an arms race, adversaries clearly have the upper hand. State of the art attacks are able to overcome most existing defense mechanisms [3], [6] and adversarial example detection algorithms [7].

Two core principles underlie of all adversarial attack methods against deep neural networks: 1) adversarial examples should closely resemble normal input, and 2) perturbation should give the adversary complete control of the classification results. In this work we propose a novel adversarial example detection method that is based on the dynamics of activation values triggered by each of a neural network classifier's layers. Our proposed detection method aims to identify behavior patterns that result from the two core principles listed above. We show that the inherent qualities of adversarial examples result in substantially different activation pattern sequences compared to normal inputs.

We evaluate our detection method using state of the art attack methods and classifier models, and show that it can effectively overcome adversarial inputs without significantly increasing computational complexity.

The reminder of the paper is organized as follows: Section II provides general background regarding adversarial machine learning, including state of the art methods for attack, defense, and detection. Section III presents our proposed detection method, which is empirically evaluated in Section **Error! Reference source not found.**. Finally, we discuss the limitations of the proposed method in Section V, and provide final notes and future research directions in Section VI.

## II. BACKGROUND

In this section we provide the background needed for understanding our detection approach. We briefly explain the existing algorithms for crafting adversarial examples, and discuss their effect over deep neural network based classifiers. We then survey known defense and detection mechanisms, in order to list the existing gaps and derive the requirements of any new detection method.

### A. Crafting Adversarial Attacks

Given a neural network based classifier $f(\cdot)$, an input vector $x$, and true class label $Y$, crafting an adversarial example $x' = (x + \delta)$ translates to identifying an adversarial perturbation $\delta$ such that

$$f(x') \neq Y \\ s.t. \|\delta\|_p < \varepsilon \quad (1)$$

where $\varepsilon$ is used to ensure the adversarial perturbation is undetectable, and $p$ is a specified distance metric. The choice of the most suitable distance metric is dependent on the classification task and context. As we deal with image classification, perturbation detectability is dependent on human perception and is not well defined using mathematical terms. However the $L_0, L_2$ and $L_\infty$ distance metrics are commonly used in this context as a proxy for human perception. $L_0$ measures the number of perturbed features within $\delta$ (e.g., image pixels), $L_2$ measures the perturbation's Euclidian norm, and $L_\infty$ measures the maximal change to any of the input features.

Various optimization methods for solving the constraints in (1) have been suggested in recent years. Typically the methods include computation of f(X), calculation of the loss gradient, and one or more steps during which the input is perturbed based on the gradient, in order to maximize the loss.

Attack methods are often divided into two classes. The approach described above is commonly known as a non-targeted attack. Here the attacker wishes to deviate from the true class label but does not care about which other class is chosen. Targeted attacks are more powerful in the sense the attacker gains full control over the classifier's output. Perturbation is constructed to assign a predetermined class label to the adversary input.

The implementation details of targeted attacks are quite similar to those of the non-targeted ones, except that in the targeted case perturbation is formed based on the gradient in the direction of the target class.

#### 1) False Sense of Certainty

In [21] the authors have created images that appear as random noise to a human observer but are actually carefully crafted adversarial examples. The optimization method used for creating those adversarial examples follows the guidelines presented in Section II.A, however instead of starting with valid input obtained from the testing set, a random starting point was used. The researchers' main contribution was in addressing the issue of model certainty. Using a state of the art ImageNet classifier, the authors showed that model certainty does not reflect reality when processing adversarial examples. Despite being fooled by the adversarial examples, the classifier was over 99% certain that the random noise images represented a valid object category.

This research highlighted a key aspect of adversarial manipulation – certainty estimations provided by deep neural network based classifiers cannot be trusted.

#### 2) Fast Gradient Sign Method and Basic Iterative Method

The fast gradient sign method (FGSM) [10] was the first computationally efficient method for crafting adversarial examples. It is a single-step attack, requiring a single iteration of gradient calculation and perturbing each of the input features by a magnitude of $\varepsilon$ in the direction of the loss gradient. Formally

$$x' = x + \varepsilon * sign(\nabla_x J(x,Y)) \quad (2)$$

where $J(x, Y)$ represent's the classifier's loss given an input vector $x$ and true class label $Y$.

The basic iterative method (BIM) [16] is an immediate extension of FGSM, where instead of performing a single step in the direction of the gradient, $h$ smaller steps of size $\gamma$ are performed so that $h * \gamma = \varepsilon$. This iterative approach allows a more refined perturbation based on the curvature of the classification manifold.

More recent research has presented effective countermeasures for both FGSM and BIM [5].

#### 3) Carlini & Wagner (C&W) Attack

In [6] the authors recently presented an attack method (commonly referred to as the C&W attack) that can overcome most, if not all, known defense mechanisms. It was specifically designed to overcome defensive distillation [26], which was considered unbeatable at the time. Their work includes a set of three attack methods (one for each commonly used distance metric), all derived from the same iterative optimization method. The result is a powerful set of attacks that are currently considered state of the art attacks for which there are no known defense or detection mechanisms.

The authors start by rephrasing the objective function used for crafting a targeted attack. Instead of requiring that

$$f(x') = t \quad (3)$$

where $t$ is the destination target class, they introduce a new objective function $q(x + \delta)$ so that

$$f(x + \delta) = t \leftrightarrow q(x + \delta) \leq 0 \quad (4)$$

The original objective function presented in (3) is highly nonlinear making it hard to solve. Using the new representation, the authors were able to construct an equivalent optimization problem that is both easier to solve and controls the tradeoff between perturbation size and the need to mislead the classifier:

$$\min \left( \|\delta\|_p + c \cdot q(x + \delta) \right) \quad (5)$$
$$s.t. \quad x + \delta \in [0,1]^n$$

By experimenting with different options for the function $q$ and the tradeoff constant $c$, the authors were able to create a highly effective attack variant for each of the three commonly used distance metrics.

*4) Black Box Attacks and Adversarial Transferability*

The research of Szegedy, C. et al. [32] pointed to the fact that adversarial examples devised using one network often cause misclassification in a different network. This is true even if the two networks have different architecture and are trained using different datasets. This remarkable finding, which was later named Adversarial Transferability [23], opened the door for black box attacks against neural networks.

Black box attacks can be launched by training a surrogate network for solving the same classification task as the target network, constructing adversarial examples against the surrogate network, and finally, for using those adversarial examples against the original target network.

Given nowadays availability (through open-source licenses) of fully trained networks for a large variety of classification tasks, black box attacks allow an adversary to launch a successful attack even when the architecture and/or weights of the target network are unknown.

*B. Defense Mechanisms*

*1) Adversarial Training*

Adversarial training is perhaps the most immediate line of defense against adversarial manipulations. It is based on iteratively training the classifier network using adversarial examples by 1) training a network to be sufficiently accurate over normal input, 2) generating adversarial examples, 3) augmenting the training input, and 4) fine-tuning the classifier.

This simple approach has demonstrated greater model resilience than undefended classifiers, however there are a few shortcomings to this approach: 1) it is difficult to scale to classifiers that process high resolution inputs like ImageNet [16], 2) adversarial training based on weak attacks does not provide an adequate defense against stronger attacks [1], and 3) it is fairly easy to construct effective adversarial examples against a network that has already been trained to cope with some adversarial examples [19].

*2) Defensive Distillation*

Distillation refers to the process of training one network over the softmax outputs of another network. Originally, this process was aimed at reducing the computational capacity associated with using a neural network. Hence, the distilled network includes a considerably lower number of neurons.

Defensive distillation [26] makes use of the same training process in order to increase the resilience of the classifier network against adversarial examples. In this case the two networks share a common size and architecture, however the distilled network is trained using a high Distillation Temperature (described below).

Distillation modifies the softmax calculation by dividing both the numerator and the denominator by the distillation temperature $T$ as follows:

$$softmax(x,T)_i = \frac{e^{x_i}/T}{\sum_j e^{x_j}/T} \quad (6)$$

With a temperature of one (6) reverts back to the standard softmax function. When distillation temperature decreases towards zero, softmax is pushed towards a "harder" max. The probability estimate of the most probable class increases towards one, while all other probabilities decrease towards zero.

As distillation temperature increases, the value of $T$ becomes much larger than $e^{x_i}$. Therefore, the softmax formula output approaches 1/N (with N denoting the number of classes). Training the distilled network using high temperature values, hence, forces it to increase the softmax input for the most probable class compared to the others. Informally we say that the classifier is required to be more certain about its classification output. Defensive distillation leverages this increased classification certainty for improving classifier resilience. It was considered a highly promising defense mechanism until Carlini and Wagner [6] designed an attack method that could effectively overcome it.

*3) Gradient Obfuscation*

As attack methods are commonly based on the calculation of the loss gradient, defense or detection algorithms should attempt to distort or eliminate the gradient calculated by an attacker. The term gradient obfuscation [24] refers to any attempt to prevent the gradient calculation, distort the gradient result, or eliminate the gradient altogether. Gradient obfuscation is considered a mandatory requirement, although not a sufficient one [3] for any defense or detection mechanism. As long as the defense mechanism is based on some differentiable function, an adversary can successfully attack both the classifier and defense models. This is done by treating the two models as a single, combined unit and computing adversarial perturbations that will fool them both [7]. Therefore, obfuscation aims to prevent such attacks.

Notable examples of obfuscation use generative adversarial networks (GANs) in an attempt to remove adversarial noise [12], [18], [27], [29]. Assuming the availability of a sufficiently accurate generator network $G(\cdot)$, those works first project the input onto the generator's latent space and then use the generator's output as input to the classifier network. Projection is performed by starting from a random point in the generator's latent space and using several steps of gradient descent to find a latent vector $z$ that minimizes $\|x - G(z)\|$. Once $z$ is found, $G(z)$ is provided as input to the classifier network and $f(G(z))$ is used as the final classification result.

Those methods are based on the assumption that no adversarial examples exist within the generator's manifold. However, in a recent work [3] the authors were able to identify adversarial examples that are close enough to the original input and placed on the generator's manifold.

Although gradient obfuscation does not ensure the resilience of a defense mechanism, it is currently considered a mandatory requirement of such mechanisms. Any defense mechanism that allows direct gradient calculation can be easily overcome. Therefore, all recently suggested methods include some non-differentiable elements, randomization, or other means of gradient elimination.

*C. Detecting Adversarial Examples*

Various attempts to detect adversarial examples based on statistical properties of the input have been suggested in recent years [34] In these studies, the authors have used various statistical tests and dimensionality reduction approaches in order to differentiate adversarial examples from normal input. However, as shown in [7], all of these approaches have failed to effectively detect state of the art attacks.

Relatively little research has attempted to use the output of inner layers of the classifier network in order to detect and defend against adversarial examples. In [35] the authors have constructed SVM classifiers trained to detect adversarial examples based on the outputs of each of the network's inner layers. They considered an input as normal, only when all of the SVM classifiers marked it as such. However, this defense method failed to defend against the more recent C&W attack, resulting in very high false positive rates [7].

The authors of [36] augmented the classifier network with a secondary detector network that is fed by the output of the classifier's convolution layers. However, given that the detector network is differentiable, more recent research [7] demonstrated that it is possible to form adversarial examples that simultaneously fool both the classifier and the detector.

Recently, Papernot, N. & McDaniel, P. [17] attempted to address the false certainty issue and ensure that the classifier yields low confidence values when faced with adversarial input. The authors constructed a set of k-nearest neighbors (k-NN) classifiers based on the output of inner convolution layers. Then, for a given input, they collected the combined list of neighbors for all k-NN classifiers and used this list in order to calculate a more accurate confidence score. High variance within this list was translated to low confidence scores and vice versa. This work differs from our approach in two key respects: 1) The authors did not construct a detector. Instead their goal was to reduce the classifier's reported certainty level when faced with adversarial input. 2) This work treats the outputs of all inner layers as a single monolithic block. The combined list of neighbors collected from all classifiers is used together for estimating the certainty score. Our approach analyzes the "spatial movements" of adversarial examples as they are processed by the different network layers. We leverage the individual class labels derived from each of the layers to form classification sequences and show that adversarial "movement patterns" reflected in those sequences are far less likely than normal ones.

III. PROPOSED SOLUTION

In this section we describe our proposed method for detecting adversarial examples. We provide implementation details and evaluation results using image classification and convolution based classifiers, however we believe the same principles can be easily adapted to other domains as well.

Our proposed solution is based on the Activation Space abstraction, a term coined by the authors to denote the hyperspace formed by the activation values of a given neural network layer.

*A. Notation and Terminology*

Building on the notation defined in Section II.A, we denote $f^i(\cdot)$ as the output of the i[th] neural network layer ($0 \leq i \leq l$). $f^0(\cdot)$ represents the network's input, while $f^l(\cdot)$ is the final softmax output. We also denote the activation space formed by the i[th] layer as $V^i$.

*B. Intuition and Motivation*

Assuming that our neural network classifier is accurate enough and thus it's input is mapped to its correct class label by the final softmax layer. Using the activation space abstraction, we say that input is mapped into distinctive clusters within $V^l$ (the activation space of the last layer).

However, in the input space, such a separation does not usually exist. The role performed by the first group of neural network layers is therefore to differentiate between classes by mapping input into a high dimensionality space, so that instances of the same class are clustered in distinct subsections. Once separation has been achieved, the final layers of the network reduce dimensionality while preserving class based separation, until the number of outputs matches the number of classes in the final layer.

Classes cannot be easily segmented within the input space, however we can assume that some classes are more easily separated than others – for instance images of dogs and cats might be harder to differentiate than those of dogs and airplanes. We can further assume that inputs that are closer in the input space require more processing (layers) in order to be separated correctly.

Adversarial examples are designed to closely resemble normal inputs but cause the classifier to assign them an incorrect class label. Within the activation spaces of the lower network layers, we therefore expect adversarial examples to appear close to normal instances of their source class. Similarly, we expect them to be in proximity to instances of some other class in the activation spaces of the last set of network layers.

All in all, we can expect different spatial behavior across various activation spaces when comparing normal and adversarial examples, and our proposed method is based on this line of thought. We track the process of class separation through the analysis of activation spaces and identify spatial patterns that differentiate adversarial examples from normal input. We believe those pattern differences are a result of the inherent nature of adversarial examples and hence expect our detection method to be resilient to future types of adversarial manipulations.

```
Algorithm 1: Detector Training
input:
    X_train  - training input set
    f(·)     - trained classifier network of l layers
    α        - maximal allowed detector false positive rate (FPR)

output:
    V        - activation spaces
    C        - k-NN classifier for each activation space
    P_s      - a priori class label switching probability
    CO       - detector's log likelihood cutoff value

logic:
    # Compute Euclidian activation spaces for each layer
    V^i ← PCA.fit(f^i(X_train))
    # Train a k-NN classifier for each activation space
    C^i ← kNN.fit(V^i(f^i(X_train)), k = 5)
    # Assign input samples with a sequence of class labels
    Ŷ^i ← C^i.predict(V^i(f^i(X_train)))

    # Calculate the a priori probability for a classification
    # change at the i^th position of a sequence
    P_s^i ← P(Ŷ^i ≠ Ŷ^{i-1}) ∀ i ∈ [1,l]

    # Construct adversarial examples
    X_adv ← C&W(X_train)

    # Calculate class switching Bayesian log likelihood
    For each x in X_train ∪ X_adv:
        LL_x ← 0
        For each layer 1 ≤ i ≤ l:
            If Ŷ^i(x) ≠ Ŷ^{i-1}(x) then
                LL_x ← LL_x + Log(P_s^i)
            Else
                LL_x ← LL_x + Log(1 - P_s^i)

    # Calculate the cutoff log likelihood value. Choose an
    # appropriate threshold value by using a ROC curve
    CO ← argmax_TPR s.t. FPR < α
```

*C. Detector Training*

The following section provides the implementation details of our detector. A pseudocode for the detector training process is provided in Algorithm 1 below.

**Constructing Activation Spaces:** During detector training we construct a baseline for modeling the behavior of normal examples within the activation spaces. We allocate half of our training set for modeling the behavior of normal, unperturbed inputs (note that the input samples used for training the detector were not used for training the classifier). We feed those inputs into the network and compute the activation values of each layer. We then construct a Euclidian activation space for each layer by projecting layer activation output using principal component analysis (PCA).

In this case the use of PCA serves a dual purpose: 1) it reduces dimensionality and in doing so removes correlated outputs and reduces the computation load, and 2) it forms a Euclidian hyperspace where axes are perpendicular to one another allowing distance calculations.

**Training a Dedicated k-NN Classifier for Each Activation Space:** Next, we train a dedicated k-NN classifier for each of the activation spaces, mapping points of that space into one of the network's class labels. It is important to note that the k-NN classification algorithm is non-differentiable. This algorithm provides us with the gradient obfuscation needed in order to block all simple, gradient based attacks against our detector.

**Estimating A Priori Class Label Switching Probability:** Based on the trained k-NN classifiers, we assign each input sample with a sequence of class labels (one label per each classifier) and further use those sequences for estimating the a priori probability for a class label change between each pair of adjacent network layers. Intuitively one can expect this a priori probability to be relatively high within the first activation spaces and to gradually decrease towards the last ones. This is a result of the improvement in class separation as we move from one network layer to another. We can also expect adversarial examples to demonstrate label switching patterns that are substantially different than those of normal ones. This is the result of the inherent goal of adversarial examples, and our experiments provide empirical evidence to support that intuition.

**Computing the Likelihood of Class Label Sequences:** Given the a priori label switching probability estimates we can compute the likelihood of a given sequence. When calculating likelihood values, we follow the naïve Bayes principle and assume that the predictions made by our k-NN classifiers are independent of one another.

We perturb the remaining half of the training samples using the C&W attack, feed the adversarial examples into the network, calculate the corresponding activation values, and ultimately assign each adversarial example with a sequence of class labels as we did for the normal examples.

Finally, we calculate the likelihood of each of the normal and adversarial sequences and choose a cutoff value to differentiate one group from the other.

**Hyper Parametrization and Implementation Notes:**
- The number of PCA components used when forming the Euclidian activation spaces affects detector accuracy. The results reported in following section are based on using the first 100 PCA components. The number of components to use was chosen through trial and error.
  For layers with more than 100 output neurons, we reduce dimensionality to 100 PCA components. For layers with fewer output neurons, we maintain the original dimensionality but use PCA to form Euclidian spaces.
- Our experiments indicate that the number of neighbors used to train the k-NN classifiers do not have a major effect on detector accuracy. We therefore report detector accuracy results using five nearest neighbors (k=5).
- Under the naïve Bayes assumption, likelihood can be computed as the multiplication product of the probabilities associated with the observed class sequences. However, when dealing with deep neural networks those sequences can grow fairly long. Multiplying the associated probabilities will therefore quickly exhaust the accuracy of floating point calculations.
  In order to overcome this computational limitation, we compute the log likelihood (the sum of probability logs) instead of the multiplication product.
- Our detector is based on estimating the log likelihood of label switching sequences assigned to normal examples and treating those as a baseline for comparison against newly provided inputs. Following anomaly detection practices by training the detector using only normal examples is aimed at increasing the efficiency of our detector against as yet unknown adversarial manipulation algorithms.

*D. Detector Evaluation*

Once the detector has been trained, evaluation is rather straight forward. We feed each new set of input images into the neural network classifier and compute the activation values for empirical evaluation. We use the trained k-NN classifiers to produce a sequence of class labels, compute the log likelihood of the observed class switches between one activation space and another, and ultimately compare the likelihood score to the cut off value.

*E. Evaluation Framework*

We conduct our experiments using the Python ecosystem. We use Keras with a TensorFlow backend as our deep learning framework, and the CleverHans [22] library for crafting adversarial perturbations.

We test our proposed method using two benchmark datasets - the MNIST handwritten digits [17] and CIFAR-10 image classification [14] datasets, as summarized in TABLE I.
For MNIST classification we use the textbook CNN described in (Keras MNIST CNN Tutorial, https://github.com/keras-team/keras/blob/master/examples/mnist_cnn.py). This classifier network includes six layers and roughly 1.2M trainable parameters, and achieves 99.1% accuracy for the unperturbed testing set. For CIFAR-10 we use a VGG16 network adjusted from (https://github.com/keras-team/keras/blob/master/examples/mnist_cnn.py). This network includes 34 layers, accounting for roughly 15M trainable parameters, and achieves 93.6% accuracy

## IV. RESULTS AND EVALUATION

*A. Evaluating the MNIST Dataset*

We follow the steps described in Section III.C, to test the validity of our approach using the MNIST dataset. We train the CNN on the predefined training set and use half of the testing samples for training our detector.

We form a targeted C&W attack [6] using the remaining 5K testing samples. A random target class is chosen for each input image.

Using the attack parameters specified in 0 we were able to manipulate the classification of all input samples to our randomly selected target class.

```
Algorithm 2: Detector Evaluation
input:
X_eval   - input for evaluation
f(·)     - trained classifier network of l layers
V        - activation spaces
C        - k-NN classifier per each activation space
P_s      - a priori class label switching probability
CO       - detector's log likelihood cutoff value

output:
Normal/Adversarial detector decision

logic:
# Compute a class label sequence for the evaluated input
For each layer 0 ≤ i ≤ l:

    # Project activation value to the iᵗʰ activation space
    PRⁱ(X_eval) ← Vⁱ.transform(fⁱ(X_eval))

    # Compute the iᵗʰ class label
    Ŷⁱ(X_eval) ← Cⁱ.predict(PRⁱ(X_eval))

# Calculate the Bayesian log likelihood of the class label
# sequence
LL_x ← 0
For each layer 1 ≤ i ≤ l:
    If Ŷⁱ(X_eval) ≠ Ŷⁱ⁻¹(X_eval) then
        LL_x ← LL_x + Log(P_sⁱ)
    Else
        LL_x ← LL_x + Log(1 − P_sⁱ)

# Compare log likelihood against the cutoff value
If LL_x < CO
    return Adversarial
Else
    return Normal
```

We start by assessing our intuition with regard to the a priori label switching probability. As shown in Fig. 1, the probability of a label switch decreases towards the final network layers, which seem to support our intuition. However, MNIST presents a rather simple classification task. This is apparent by the relatively low switching probability for the first layers, as well as by this simple model's near perfect classification result.

With our intuition affirmed, we move on to calculating the log likelihood of class switching for normal and adversarial examples. As is clear from Fig. 1, the likelihood of the adversarial switching sequences is considerably lower than that of normal ones. This is a result of the inherent nature of adversarial examples. Namely, they force a class switch in the last layers, where the a priori probability for a switch is very low.

The last step for constructing the detector is choosing a likelihood threshold value. Fig. 2 provides a sensitivity analysis of different threshold values using an ROC curve. In this context, true positive refers to correctly identified adversarial examples, while false positive refers to cases of normal images that were incorrectly classified as adversarial. As is illustrated in Fig. 2, our detector achieves an AUC of 0.91 when applied to all testing instances.

*B. Evaluating the CIFAR-10 Dataset*

The MNIST dataset cannot prove the validity of a defense mechanism by itself [7]. Some defense mechanisms have been shown to be efficient in blocking/detecting adversarial perturbations against the MNIST dataset, but they failed to operate on more complex use cases. Therefore, we repeated the experiment using the CIFAR-10 dataset.

The increased complexity of the CIFAR-10 classification task is reflected in the inner structure of the CNN required to solve it: there are many more layers and over 10 times more trainable parameters, compared to the MNIST CNN. As in the case of MNIST, we used a randomly selected target for each input sample. The C&W attack parameters utilized in this case are listed in 0It is worth noting that we employ a much stronger attack configuration in this case, with more iterations, greater attack confidence, and a lower learning rate. Using this configuration, we were able to achieve 100% success in manipulating classification to our selected target class.

TABLE I. DATASETS, CLASSIFIER NETWORKS, AND ACCURACY FIGURES

| Dataset | Accuracy | Input | Train/Test Images | Network Architecture |
|---|---|---|---|---|
| **MNIST** | 99.1% | 28x28 grayscale images | 60K / 10K | Textbook solution (6 layers, 1.2M parameters) |
| **CIFAR-10** | 93.6% | 32x32 RGB images | 50K / 10K | VGG16 (34 layers, 15M parameters) |

TABLE II. C&W ATTACK PARAMETERS FOR THE MNIST AND CIFAR-10 EXPERIMENTS

| Attack Parameter | MNIST | CIFAR-10 |
|---|---|---|
| **Iteration Count** | 200 | 1000 |
| **Learning Rate** | 0.1 | 0.01 |
| **Initial Constant (for Binary Search)** | 10 | 0.001 |
| **Confidence** | 0 | 5 |
| **Binary Search Steps** | 1 | 9 |

As a first analysis step, we provide a visual analysis of the expressive power of activation spaces. We projected each of the activation spaces of the CIFAR-10 classifier network onto a 2D plot using t-Distributed Stochastic Neighbor Embedding (t-SNE) [38]. t-SNE is an unsupervised dimensionality reduction method that is particularly suitable for visualizing high dimensionality datasets. This method attempts to preserve, within the projected visualization, the relative distance from each data point to its neighbors. As a result, the distance between every two data points on the t-SNE visualization is correlative to their distance in the original dataset. It should be noted, however, that the absolute position within the visualization is meaningless.

Fig. 3. includes The t-SNE visualization of every fourth activation space in the CIFAR-10 classifier. The colors represent different true class labels. Notably, the t-SNE algorithm itself is unsupervised. The true class labels are only used for visualization and are not used as part of the t-SNE projection process. The gradual increase in class separation is clearly evident, supporting our initial intuition. Starting from an intertwined mixture of colors in the t-SNE visualization of the first activation space, and ending with distinctive color areas in the t-SNE visualization of the last activation space.

Next we have tested the performance of our detector over the CIFAR-10 dataset. The results of this experiment are presented in Fig. 4. As we anticipated, a priori label switching probability decreases towards the last layers of the network. In contrast, adversarial switching probability remains much higher through most of the network's layers. Those vastly different switching patterns yield substantially lower log likelihood values for the adversarial examples.

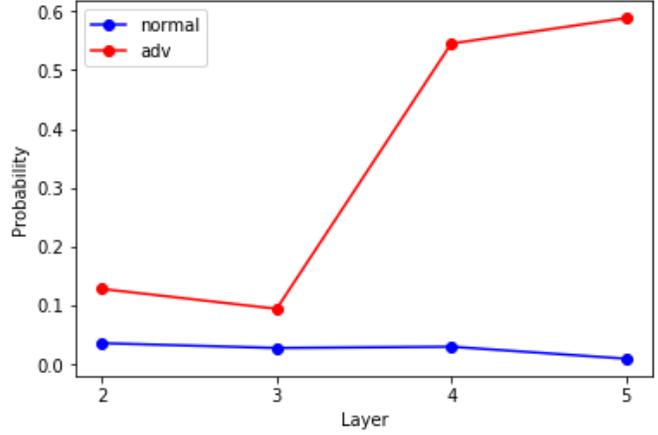

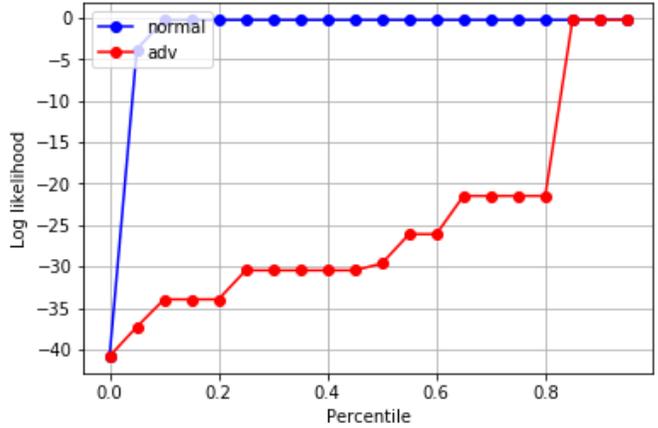

Fig. 1. Results on the MNIST dataset. Top - class switching probability for normal and adversarial input; bottom - log likelihood of class switching sequences.

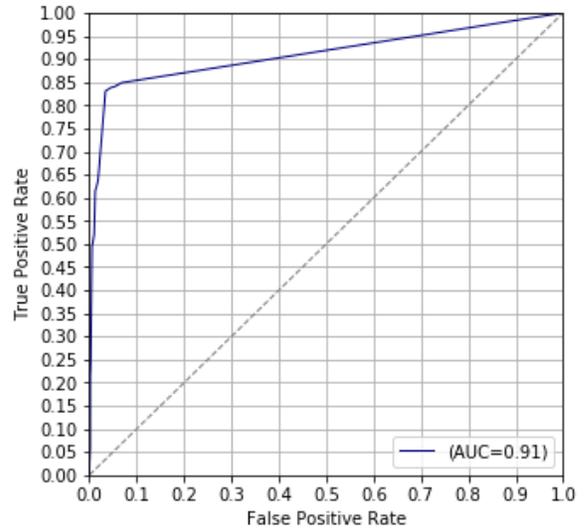

Fig. 2. Detector ROC curve for the MNIST dataset.

Testing the correlation between the calculated log likelihood and the perturbation norm showed that the two are practically independent of one another (Pearson correlation of 0.009). This supports our claim that our detector is based on an inherent quality of adversarial examples, as opposed to some visual artifact, and increases our confidence regarding the validity of our method against future attacks.

As previously mentioned, the last step for constructing the detector is choosing a likelihood threshold value. Fig. 5 provides a sensitivity analysis of different threshold values using an ROC curve. Our detector achieves an AUC of 0.92 when applied to all testing instances.

An in depth investigation of the false positive cases reveals that they are largely associated with cases of misclassification by the original classifier network. Whereas on average our network provides incorrect classification for 6.3% of all examples, it misclassifies roughly 30-40% of all false positive examples (depending on the threshold value).

We see that incorrectly classified images exhibit a large number of label switches and a high rate of switches in the last 10 layers of the network, similar to adversarial examples. This further strengthens our understanding of the role played by the different layers of the DNN.

Replotting the ROC using only correctly classified examples shows an impressive AUC of 0.95.

## V. LIMITATIONS OF THE PROPOSED FRAMEWORK

While our proposed method only marginally increases the computation load of the original classifier, its memory requirements are proportional to the depth of the original classifier network. In the case of modern convolutional networks (e.g., ResNet) these memory requirements can increase significantly.

Our detector uses a set of k-NN classifiers as a form of gradient obfuscation. Ideally, we would like to have an all DNN solution making the detector an integral part of the classifier

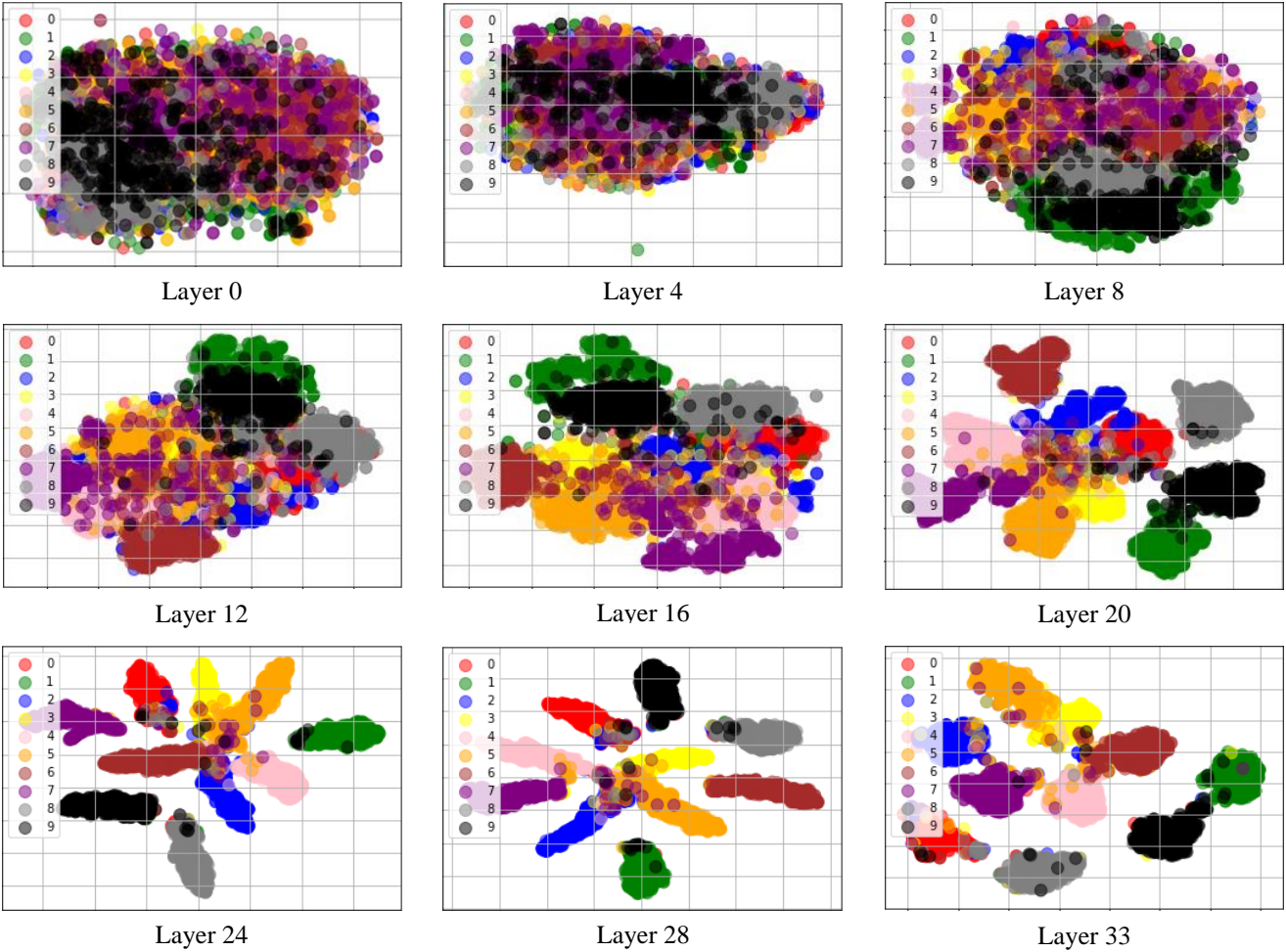

Fig. 3. t-SNE based visualization of the CIFAR-10 classifier activation spaces

## VI. SUMMARY

In this work we present a novel method for detecting adversarial perturbations based on activation spaces, the hyperspaces formed by the activation outputs of the network's inner layers. We demonstrate the potential value of our approach using two commonly used verification datasets (MNIST and CIFAR-10) and C&W, the state of the art attack method [6]. Our proposed detector achieves an AUC of 0.95 on the CIFAR-10 dataset.

We designed our detector taking into account the inherent nature of adversarial examples. Our experiments provide initial evidence that we were able to detect spatial behavior in activation spaces that is related to this nature. Our proposed detector leverages a set of k-NN classifiers trained on the activation outputs of each layer of the neural network. We measure the a priori probability of class label changes between every two consecutive layers and use this set of probabilities to compute the likelihood of the entire classification sequence. The k-NN classification algorithm is non-differentiable, hence preventing simple attacks against our detector model.

In an attempt to make our algorithm future ready for handling as yet unknown attack methods, we compute our likelihood baseline estimates using only normal, unperturbed input.

Intuitively, we hypothesized that DNN classifiers begin by clustering input samples in a high dimensional space and then reduce representation dimensionality towards the final layer, while preserving class separation. Our experiments support this intuition in two ways: 1) the a priori label changing probability decreases asymptotically towards the final layers of the network, and 2) false positive detection is tightly coupled with classification accuracy. Normal inputs that are misclassified by the original neural network are very likely to trigger a false positive classification by our detector model.

In the future we plan to extend our approach to include additional network types such as RNNs, by leveraging of the CNN to RNN transferability. We would like to explore ways of using an all DNN implementation of our proposed method in an attempt to start bridging the gap between human perception and DNN based classification. We also plan to continue to explore the nature of adversarial examples in different content domains with various classifier architectures.

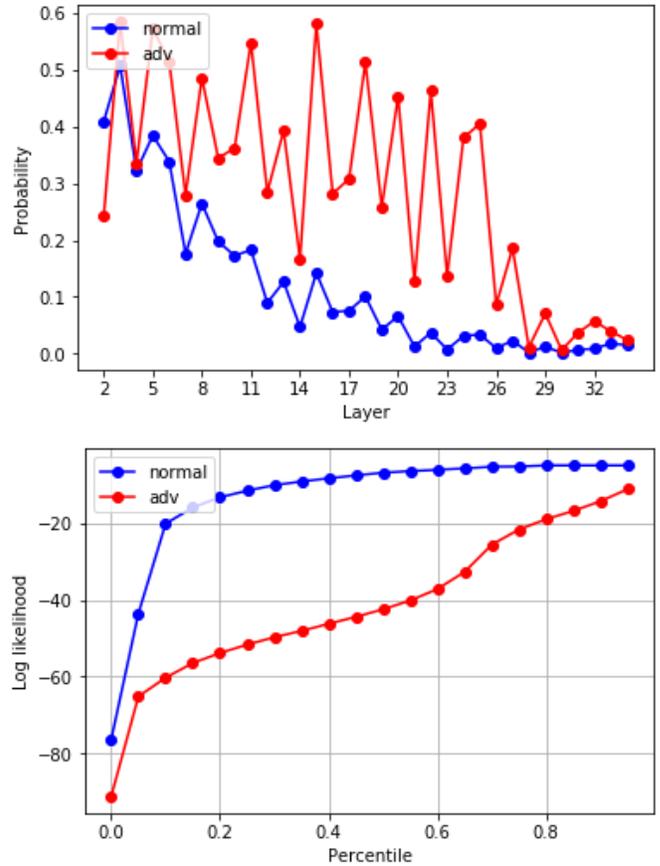

Fig. 4. Results on the CIFAR-10 dataset. Top - class switching probability for normal and adversarial inputs; Bottom - log likelihood of class switching sequences.

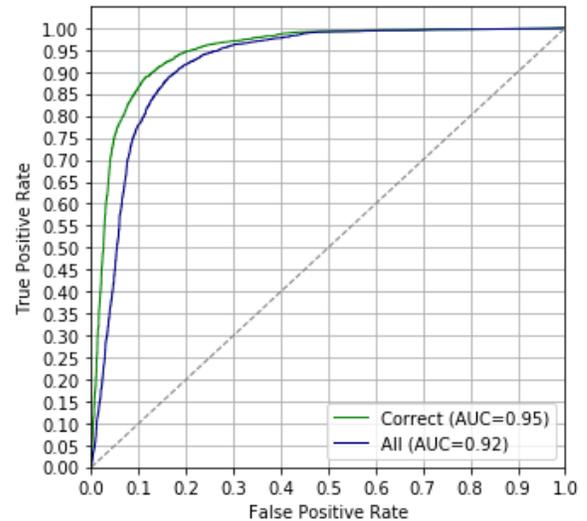

Fig. 5. Detector ROC curve on the CIFAR-10 dataset. Green represents the results filtered for inputs that are correctly classified by the CIFAR-10 network. Blue represents all testing inputs without filtering.